# CONVEXITY ANALYSIS OF SNAKE MODELS BASED ON HAMILTONIAN FORMULATION

GILSON A. GIRALDI[1]
ANTONIO ALBERTO FERNADES DE OLIVEIRA[1]

[1]LCG-Computer Graphics Laboratory, COPPE-Sistemas, UFRJ
21945-970, Rio de Janeiro, RJ, Brazil, Caixa postal 68511
{giraldi,oliveira}@cos.ufrj.br

**Abstract.** This paper presents a convexity analysis for the dynamic snake model based on the Potential Energy functional and the Hamiltonian formulation of the classical mechanics. First we see the snake model as a dynamical system whose singular points are the borders we seek. Next we show that a necessary condition for a singular point to be an attractor is that the energy functional is strictly convex in a neighborhood of it, that means, if the singular point is a local minimum of the potential energy. As a consequence of this analysis, a local expression relating the dynamic parameters and the rate of convergence arises. Such results link the convexity analysis of the potential energy and the dynamic snake model and point forward to the necessity of a physical quantity whose convexity analysis is related to the dynamic and which incorporate the velocity space. Such a quantity is exactly the (conservative) Hamiltonian of the system.

## 1. Introduction

This paper addresses some issues concerned the convexity analysis and dynamic evolution for the class of "deformable models" originated with the method of "snakes" [Kass at al. (1987)] which are used to locate boundaries in 2-D and 3-D imagery and for tracking [Black-Yuille (1992)].

These models are known as "Snakes" or "Active Contour Models (ACM)" and can be formulated by the minimization of an energy functional that embodies the image-based information.

The main advantage of snakes is that they are topologically isomorphic to the features they seek, namely object boundaries. As a result, no edge linking is required, and they are robust to low contrast, noise, and gaps or spurious branches in boundaries [Black-Yuille (1992)].

The main disadvantage is that their convexity properties are not clear. Specifically, it has been noted in the past that active contour models are nonconvex, and that solutions are often locally rather globally optimal solutions, often involving discontinuities or "splits" in the final contour [Davatzikos-Prince (1996)].

The original proposal to bypass this problem is embedding the curve in a viscous medium and solving the dynamics equation resulting Black-Yuille (1992)]. This methodology has the advantage of allowing the initial curve to pass local minima and to stop only in a global one in a region of interest. However, the performance and efficiency of the model depends on new dynamic parameters (mass $\mu$ and viscous damping $\gamma$) and initial conditions (position and velocity), which can not be determined a priory [Leymarie-Levine (1993)]. For example, if the initial velocity is not chose properly, the snake can stops in a position that is far from the border we seek.

Levine at al. [Leymarie-Levine (1993)] used another approach by applying hierarchical filtering methods, as well as a continuation method based on a discrete scale-space representation. Basically, a scale-space scheme is first used at a coarse scale to get closer to the global energy minimum represented by the desired contour. In further steps, the optimal valley or contour is sought at increasingly finer scales.

Another approach, the Dual Active Contour Model [Gunn-Nixon (1997)], try to go away from local minima by using two contours: one contour which expands from inside the target feature, and another one which contracts from the outside. The two contours are interlinked to provide a driving force to carry the contours out of local minima. Despite this ability, this approach can treat only simple shapes and its extension to 3D is too expensive.

Another approach to deal with the convexity problem is do search for conditions to guarantee the convexity for the energy functional in a region $R'$ of interest. In [Davatzikos-Prince (1996)] we find such a methodology based on a discrete snake model. Having the guarantee of convexity a optimization method can be applied with sure of convergence.

The result of such methodology can be applied, at first, to any kind of external force with the computational cost of finding a lower bound for the minimum eigenvalue of the Hessian of the energy as we will show later.

In this work we will suppose that the curve can be represented as a point in a finite dimensional space (*configuration space*) that we generically call *Q space*. Related to it, in a way that we shall clarify later, is the *velocity space* that we call $\dot{Q}$ *space*.

The main proposal of this work is to extend the convexity analysis of the energy functional presented by Prince and Davatzikos [Davatzikos-Prince (1996)] in a way to incorporate the velocity space. The main consequence of this analysis is to find conditions to guarantee the convergence of the solution of the dynamic snake model to the desired contour.

In this way, we first analyze the dynamic snake model as a dynamic system to clarify the role of the extremes of the potential energy in the space of positions and velocities correspondent. The result of this analysis is that the extremes of the potential energy are the singular points of that system.

The next steep is naturally to study the singular points to classify them in accordance with the eigenvalues of the Hessian of the energy. The main conclusion is that if the border is a local minimum of the potential energy and we do not have mass negative then the border will be an attractor in that space.

So, this analysis implies that there is a region $V$ in the $(Q, \dot{Q})$ space such that any initial condition $(Q_0, \dot{Q}_0) \in V$ is attracted to the point (border) we seek.

To define this region it is necessary to extend the convexity analysis of the energy in a way to incorporate the velocity space. This can be accomplished in a natural way by using the Hamiltonian formulation [Goldstein (1980)] of the snake model.

In fact *the Hamilton's Equations for the snake model* gives a framework to link the dynamic features of the snake model and the convexity analysis of the energy functional presented in [Davatzikos-Prince (1996)].

In the following sections we review the convexity analysis for snakes given in [Davatzikos-Prince (1996)]. Section 3 describes the snake model of interest. The numerical analysis correspondent is given in section 4 and a qualitative analysis of the system in section 5.

Next, the *Hamiltonian Formulation* is presented and discussed. Finally, we present our conclusions and future directions for this work.

## 2. Active Contour Models

Let $(Ad, E)$ be a space of curves:

$$c:[0,1] \to D \subset \Re^2, \quad c(s) = (x(s), y(s)), \quad (2.1)$$

where D is a domain of interest that can be the all domain of the image or only a part of it and $c \in C^4$.

Let $E: Ad \to \Re$ a functional, which we want to extremize. The par $(Ad, E)$ defines an Deformable Model [Cohen (1991)].

In the Active Contour Models (*ACM*), the functional $E$ is constructed considering the curve a thin string immersed in a external potential *P* derived from the image and subjected to internal forces due elastic properties: elasticity $(\omega_1)$ and rigidity $(\omega_2)$.

Using a kind of *thin-plate-under-tension* model [Kass at al. (1987)] for the elastic energy, we can express the Potential Energy Functional of the model by:

$$E_p(c) = \int_0^1 \left[ \omega_1 \left\| \frac{dc}{ds} \right\|^2 + \omega_2 \left\| \frac{d^2 c}{ds^2} \right\|^2 \right] ds + \int_0^1 P(c) ds.$$

$$(2.2)$$

It can be shown by using the calculus of variations that a curve passing through two points $P_1$ and $P_2$ with known tangents $c'(0)$ and $c'(1)$ at these points and that minimizing $E_p$ must satisfy the Euler-Lagrange Equations[Cohen (1991)]:

$$-2\frac{d}{ds}\left(\omega_1 \frac{dc}{ds}\right) + 2\frac{d^2}{ds^2}\left(\omega_2 \frac{d^2 c}{ds^2}\right) - \vec{\nabla} P = 0.$$

$$(2.3)$$

The convexity analysis of this model could be done by the second variation of the functional $E_p$ [Dubrovin at al. (1984)]. Unfortunately, the results for such analysis are not practical which have driven the research in the direction of discrete versions of the model (2.2). With such simplification, the functional $E_p$ becomes a nonlinear function and the ACM becomes a nonlinear optimization problem.

Specifically, let's consider the active contour as a collection of points

$$\{q_i = (x_i, y_i)^T, i = 0, 1, \ldots, N\}, \quad (2.4)$$

where $q_i = c(i/N)$. If we have the points $q_0$ and $q_N$ fixed (open contour), the vector of the free values of the discrete curve is:

$$d = \begin{pmatrix} q_1 & q_2 & \ldots & q_{N-2} & q_{N-1} \end{pmatrix}^T \quad (2.5).$$

Then, we obtain the discrete approximation of the energy functional $E_p$ by applying finite difference approximation of the derivatives in (2.2):

$$\tilde{E}(d) = \tilde{E}_e(d) + \tilde{E}_c(d), \quad (2.6)$$

where the exact expressions for $\tilde{E}_e(d)$ and $\tilde{E}_c(d)$ can be found in [Davatzikos-Prince (1996)].

The necessary condition for $d$ to minimize $\tilde{E}(d)$ is:

$$\nabla \tilde{E}(d) =$$
$$\omega_1 N(2A_1 d - b_1) + \omega_2 N^3 (2A_2 d - b_2) + \nabla_d P = 0, \quad (2.7)$$

where $b_1$ and $b_2$ are $2(N-1)$ vectors related to the boundary conditions and $A_1 = diag(B_1)$ and $A_2 = diag(B_2)$ where $B_1$ is a symmetric Toeplitz tridiagonal matrix whose first row is $(2, -1, \ldots, 0)$ and $B_2$ is a symmetric tridiagonal matrix (see [Davatzikos-Prince (1996)] for details). If $\tilde{E}(d)$ is strictly convex then (2.7) is also a sufficient condition, and $d$ is the unique minimizer of $\tilde{E}(d)$.

In what follows, the domain where the potential $P$ is defined is denoted by $R$. The domain $D$ in which $d$ is defined is then given by:

$$D = \left\{ \begin{array}{l} (e_1, e_2, \ldots, e_{n-1})^T \in \Re^{2N-2}; e_i \in R, \\ i = 1, \ldots, N-1 \end{array} \right\}. \quad (2.8)$$

**2.1 Convexity Analysis**

The point here is to find conditions for which $\tilde{E}(d)$ is strictly convex. In [Davatzikos-Prince (1996)] we such analysis which we summarize here. The fact that $\tilde{E}(d)$ is a scalar function defined in $D$ implies that $\tilde{E}(d)$ will be strictly convex in a region $R$ if the Hessian matrix $D^2 \tilde{E}(d)$ of $\tilde{E}(d)$ is positive defined in $R$; that is, if the smallest eigenvalue of $D^2 \tilde{E}(d)$ is greater than zero:

$$\lambda_{min}(D^2 \tilde{E}) > 0 \Leftrightarrow \lambda_{min}(D^2 \tilde{E}_e + D^2 \tilde{E}_c) > 0$$

where $D^2 \tilde{E}_e(d)$ and $D^2 \tilde{E}_c(d)$ are the Hessian matrix of $\tilde{E}_e(d)$ and $\tilde{E}_c(d)$, respectively.

Using the fact that the smallest eigenvalue of the sum of two symmetric matrices is greater than or equal the sum of the smallest eigenvalues of the two matrices, we have the following sufficient condition: $\tilde{E}(d)$ is strictly convex if

$$\lambda_{min}(D^2 \tilde{E}_e) + \lambda_{min}(D^2 \tilde{E}_c) > 0. \quad (2.9)$$

It can be shown that the eigenvalues of $D^2 \tilde{E}_e$ coincide with those of the matrix $2\omega_1 N A_1 + 2\omega_2 N^3 A_2$ [Davatzikos-Prince (1996)].

Moreover, since $A_1$ is block diagonal, its eigenvalues coincide with those $B_1$. Similarly, the eigenvalues of $A_2$ coincide with those of $B_2$. Therefore, we conclude that

$$\lambda_{min}(D^2 \tilde{E}_e) \geq 2\omega_1 N \lambda_{min}(B_1) + 2\omega_2 N^3 \lambda_{min}(B_2) > 0. \quad (2.10)$$

Since $B_1$ is Toeplitz tri-diagonal matrix, its eigenvalues can be found through a recursion formula [Grenander-Szego (1958)], given the following minimum eigenvalue:

$$\lambda_{min}(B_1) = 2(1 - \cos(\pi/N)). \quad (2.11)$$

The eigenvalues of $B_2$ satisfy a double recursion formula which does not have a explicit solution but have a lower bound given by:

$$\lambda_{min}(B_2) \geq (\lambda_{min}(B_1))^2. \quad (2.12)$$

Substituting (211) and (2.12) into (2.10) we find the result

$$\lambda_{min}(D^2 \tilde{E}_e) \geq$$
$$4\omega_1 N(1 - \cos(\pi/N)) + 8\omega_2 N^3 (1 - \cos(\pi/N))^2 \quad (2.13)$$

## 2.2 Minimum Eigenvalue of $D^2 \tilde{E}_c$

First we have to notice that the matrix $D^2 \tilde{E}_c$ is a block diagonal with diagonal entries given by:

$$D_i = \frac{1}{N} \begin{pmatrix} \dfrac{\partial^2 P(q_i)}{\partial x^2} & \dfrac{\partial^2 P(q_i)}{\partial x \partial y} \\ \dfrac{\partial^2 P(q_i)}{\partial x \partial y} & \dfrac{\partial^2 P(q_i)}{\partial y^2} \end{pmatrix}, i = 1, ..., N-1$$

(2.14)

where $q_i = (x_i, y_i)^T$. Thus, the eigenvalues of $D^2 \tilde{E}_c(d)$ can be determined by finding the eigenvalues of these 2x2 matrices. A direct solution yields:

$$\lambda_{i1} = \left(P_{xx}(p_i) + P_{yy}(p_i)\right)/2N + \sqrt{\left(P_{xx}(p_i) - P_{yy}(p_i)\right)^2 + 4P_{xy}^2(p_i)}/2N, \quad (2.15)$$

$$\lambda_{i2} = \left(P_{xx}(p_i) + P_{yy}(p_i)\right)/2N - \sqrt{\left(P_{xx}(p_i) - P_{yy}(p_i)\right)^2 + 4P_{xy}^2(p_i)}/2N, \quad (2.16)$$

for $i=1,...,N-1$. The minimum eigenvalue is:

$$\lambda_{min}\left(D^2 \tilde{E}(d)\right) = min\{\lambda_{i1}, \lambda_{i2}, i = 1, ..., N-1\}.$$

We can go a step further to eliminate the dependence of these expressions on the specific location of the points $q_i = (x_i, y_i)^T$. To do this we define:

$$h_1 = \left(P_{xx}(x,y) + P_{yy}(x,y)\right)/2N + \sqrt{\left(P_{xx}(x,y) - P_{yy}(x,y)\right)^2 + 4P_{xy}^2(x,y)}/2N,$$

(2.17)

$$h_2 = \left(P_{xx}(x,y) + P_{yy}(x,y)\right)/2N - \sqrt{\left(P_{xx}(x,y) - P_{yy}(x,y)\right)^2 + 4P_{xy}^2(x,y)}/2N,$$

(2.18)

Let's see how it is possible to simplify this expressions. First we consider the isopotential curve passing through a point $(x,y)^T \in R$ and assume at first that its curvature at $(x,y)^T$ is nonzero. We can then define a local polar coordinate system whose origin coincides with the center of curvature of this curve (see [Xu at al. (1994)] for details). We now view the potential as a function of $r$ and $\phi$, the radius and angle in the local coordinate system, and seek expressions for $P_{xx}$, $P_{yy}$, and $P_{xy}$ at $(x,y)^T$.

First, it can be shown that $P_\phi = P_{\phi\phi} = 0$ [Davatzikos-Prince (1996)]. Second, we note that the potentials used for locating boundaries have isopotential curves that are nearly parallel to each other, and that are nearly parallel to the boundary (this is specially true when the boundaries are smooth or the point $(x,y)^T$ is very near the desired boundary).

By this observation it is possible to set $P_{r\phi} \approx 0$, and finally, to approximate (2.17)-(2.18) by [Davatzikos-Prince (1996)]:

$$h_1 \approx$$

$$\frac{1}{4}\left(P_{rr}(x,y) + \frac{P_r(x,y)}{r(x,y)}\right) + \left|P_{rr}(x,y) - \frac{P_r(x,y)}{r(x,y)}\right|, (2.19)$$

$$h_2 \approx$$

$$\frac{1}{4}\left(P_{rr}(x,y) + \frac{P_r(x,y)}{r(x,y)}\right) - \left|P_{rr}(x,y) - \frac{P_r(x,y)}{r(x,y)}\right|, (2.20)$$

Therefore, it is straightforward to show that:

$$\min_{x \in (x,y)}\{h_1(x,y), h_2(x,y)\} = \min_{x \in (x,y)}\{e_1(x,y), e_2(x,y)\}$$

where

$$e_1(x,y) = \frac{P_{rr}(x,y)}{2} \text{ and } e_2(x,y) = \frac{P_r(x,y)}{2r(x,y)}$$

By using this and the fact that

$$\lim_{N \to \infty} 2N^2\left(1 - \cos(\pi/N)\right) = \pi^2,$$

we can eliminate the dependence of $N$ in (2.9) and finally put the final convexity condition in the following form:

$$A(R') + \omega_1 \pi^2 + \omega_2 \pi^4 > 0, \quad (2.21)$$

where $A(R') = \min_{x \in R'}\{e_1(x,y), e_2(x,y)\}$ and $R'$ is a subset of $R$ which contains the solution we seek.

With such analysis it is possible to garantee that an optimization method will converge to the solution of (2.7) in $R'$.

Another approach to optimize the energy $E_p$ is to embed the curve in a viscous medium and solving the dynamics equation corresponding [Black-Yuille (1992)]. This approach can be more robust against local minima although it is still sensible to the poor convexity of $E_p$. In the following sections we present a formulation for this approach.

### 3. Dynamic Snake Model

In this case, the idea is to submit the initial curve to a Newtonian dynamics generated by an external potential $P$, internal (elastic) force and a viscous medium that dissipates kinetic energy.

In this discussion we are supposing deformable curves that can be represented in a form:

$$c(s) = \sum_i q_i f_i(s) = B^T Q, \quad (3.1)$$

where $B = \{f_0, f_1, ..., f_N\}$ is a set of linearly independent functions and $Q = (q_1, q_2, ..., q_N)^T$ is a column vector of points of the plane which are called the *control points* of the curve.

The most common examples of these deformable models are the D-NURBS [Qin-Terzopoulos (1996)] and Active Splines [Black-Yuille (1992)] which based on piecewise polynomial functions belonging to the space of measurable, square-integrable function denoted by $L_2$ [Bartels at al. (1987)].

The instantaneous configuration of the system is given by a point in a *configuration space* of dimension *(2N+2)*, whose coordinates are given by the *N+1* generalized $q_i$ in the $Q$ vector above. As the curve is deformed, the control points get move, so we have a velocity vector $\dot{Q}$ associate with $Q$. The complete state of the system (snake) is defined by a pair $(Q, \dot{Q})$.

An advantage of using curves of the form (3.1) is that the motion equations in the $Q$ space (Euler-Lagrange equations) are ordinary differential equations.

To see this, let's take the energies and dissipation (damping) term given by:

**Kinetic Energy**

$$T(\dot{c}) = \frac{1}{2}\int \mu \left\|\frac{\partial c}{\partial t}\right\|^2 ds =$$

$$\frac{\mu}{2}\dot{Q}^T M_0 \dot{Q} \quad (3.2)$$

where $M_0 = \int BB^T ds$ and $\mu$ is de linear mass.

**Dissipation**

$$\frac{\partial U_{at}}{\partial t}(\dot{c}) = -\frac{\gamma}{2}\dot{Q}^T M_0 \dot{Q} \quad (3.3)$$

where $\gamma$ is the constant damping density.

**Elastic Potential Energy**

$$E_e = \omega_1 Q^T M_1 Q + \omega_2 Q^T M_2 Q, \quad (3.4)$$

where

$$M_1 = \int \frac{dB}{ds} \cdot \frac{dB^T}{ds} ds, \quad (3.5)$$

$$M_2 = \int \frac{d^2 B}{ds^2} \cdot \frac{d^2 B^T}{ds^2} ds, \quad (3.6)$$

**Field Potential Energy**

$$E_c = \int P(c(s,t)) ds. \quad (3.7)$$

**Potential Energy**

$$E_p = E_e + E_c. \quad (3.8)$$

The Euler-Lagrange Equations derived from the Hamilton's Principle [Goldstein (1980)] for the system described by (3.2)-(3.8) compose the snake model that we will study here and the curve that moves to seek the solution we call *snake*.

The Euler-Lagrange Equations for the snake have the form [Curwen-Blake (1993)]:

$$\frac{d}{dt}\left(\frac{\partial T}{\partial \dot{Q}}\right) - \frac{\partial T}{\partial Q} - \frac{\partial}{\partial \dot{Q}}\left(\frac{\partial U_{at}}{\partial t}\right) + \frac{\partial E_p}{\partial Q} = 0. \quad (3.9)$$

By using (3.2)-(3.8) we find the following set of equations:

$$\mu \ddot{Q} + \gamma \dot{Q} + M_0^{-1}(\omega_1 M_1 + \omega_2 M_2)Q = M_0^{-1} F. \quad (3.10)$$

where:

$$F = -\frac{1}{2}\int \frac{\partial P(c(s,t))}{\partial Q} ds. \quad (3.11)$$

The parameters $(\mu, \gamma, \omega_1, \omega_2)$ must be chosen in advance and are essential for the behavior of the model. We have also to constrain this system to initial conditions to have a unique solution.

**3.1 Parameters Analysis**

Before continuing it will be useful to analyze some features related to the parameters of the model.

**3.1.1 Tension and Rigidity**

The effects of these parameters in the behavior of the model are studied in works like [Leymarie-Levine (1993)]. Despite this, the development of systematic procedures to estimate them has received little attention [Fisker-Carstensen (1998)],

The convexity analysis presented in [Davatzikos-Prince (1996)] is a way to limit the range for these parameters. Another possibility would be to allow these parameters to vary in time or space [Leymarie-Levine (1993)] which increases the computational cost.

In [Xu at al. (1994)] we find another approach based on adding a new energy term such that the normal force exerted at every point is equal regardless of contour shape. The new model is less sensible to the parameters but the computational cost is increased and the extension to 3D is difficult.

In this work, we will constrain tension ($\omega_1$) and rigidity ($\omega_2$) to the convexity analysis of the potential energy. So, the methodology is to keep them constants that might be updated during the snake evolution. Such proposal is also interesting for saving time calculation.

**3.1.2 Mass and Damping Densities**

These parameters are important to control the stability of the numerical method and the rate of convergence corresponding as we show below.

The ideal situation is to have these parameters such that the velocity of the snake decays very fast only when it is close the border we seek.

Such behavior is known as "critical damping" in the theory of oscillations [Nayfeh-Mook (1979)] and means that the transient part of the solution decays faster.

In a model described by (3.10) we have to be careful because it is in general a nonlinear system. For example if we simplify the energy (3.8) in a way that the equation of motion becomes that of a forced harmonic oscillator [Curwen-Blake (1993)], we have "critical damping" in some *modes* (defined next) but do not have in another ones. The same is true for the general model (3.10) as we show next. Now, let's see the effects of these parameters in the convergence of the numerical method.

**3.1.3 Numerical Method**

The equation (3.10) do not have in general analytical solution so we have to use a numerical approach to solve it with the desired precision.

The traditional methodology found in the literature [Blake-Yuille (1992)] is to use finite difference methods (FDM) in time and space. Another possibility is to apply finite differences in time but finite element methods (FEM) in space [Cohen-Cohen (1993)].

Following [Lemayre-Levine (1993)] we have chose FDM for time and space because of its simplicity and because its convergence and numerical stability properties seem to be adequate as we show now. However, for Active Surface Models, the FEM is probably a better choice [Cohen-Cohen (1993)].

So, let's present the numerical scheme used and how the dynamic parameters $\mu$ and $\gamma$ control the behavior of it.

The finite differences scheme we use is given by the following approximations of the first and second time derivatives [Chapra-Canale (1988)]:

$$\dot{Q}(t) \approx \frac{1}{2\tau}[Q(s,t) - Q(s, t-2\tau)], \quad (3.12)$$

$$\ddot{Q}(t) \approx \frac{1}{\tau^2}[Q(s,t) - 2Q(s, t-\tau) + Q(s, t-2\tau)], \quad (3.13)$$

with analogous expressions for space derivatives.

Through these expressions we can discretise the equation (3.10) in the form:

$$[(2\mu + \gamma\tau)M_0 + 2K\tau^2]Q^{t+\tau} =$$
$$2\tau^2 F(t-\tau) + 4\mu M_0 Q^t - (2\mu - \tau\gamma)M_0 Q^{t-\tau}, \quad (3.14)$$

where $Q^t$ and $Q^{t-\tau}$ are given in advance and

$$K = \omega_1 M_1 + \omega_2 M_2, \quad (3.15).$$

By dividing the two sides of (3.14) by $2\tau^2$ we find the above linear system which we have to solve in each interaction:

$$AQ^{t+\tau} = B^{t, t-\tau}. \quad (3.16)$$

where:

$$A = \left[\left(\frac{\mu}{\tau^2} + \frac{\gamma}{2\tau}\right)M_0 + K\right], \quad (3.17)$$

$$B = 2F(t-\tau) + \frac{2\mu}{\tau^2}M_0 Q^t - \left(\frac{\mu}{\tau^2} - \frac{\gamma}{2\tau}\right)M_0 Q^{t-\tau}, \quad (3.18)$$

Let's now study the properties of this scheme. For a numerical scheme gives coherent results it has to have the properties of consistence, stability and convergence [Hirsch (1988)].

The consistence is straightforward verified due the expressions (3.12)-(3.13).

For the stability condition of the scheme let's first suppose $D_2 F$ limited. So, the analysis of the stability can be carried out by the conditioning of matrix $A$. If $A$ is well-conditioned then we have a guarantee of stability for the scheme.

Following the methodology found in [Leymarie-Levine (1993)], let's take the condition number of the matrix $A$ in (3.17). Supposing $A$ is non singular, this number is given in 2-norm by [Golub-Loan(1985)]:

$$\kappa_2(A) = \frac{\lambda_{\max}(A)}{\lambda_{\min}(A)}, \quad (3.19)$$

where

$\lambda_{\min}(A) = \min_i(|\lambda_i(A)|)$, $\lambda_{\max}(A) = \max_i(|\lambda_i(A)|)$, and $\lambda_i(A)$ indicate the eigenvalues of $A$.

Through the sensibility analysis for linear systems we know that the closer $\kappa_2(A)$ is to unity, the better is the conditioning of $A$ [Golub-Loan(1985)].

We can notice $A$ has the form $A = \beta M_0 + K$ and so:

$$\kappa_2(A) = \kappa_2(\beta M_0 + K) \leq \frac{\beta \lambda_{\max}(M_0) + \lambda_{\max}(K)}{\beta \lambda_{\min}(M_0) + \lambda_{\min}(K)}, \quad (3.20)$$

which can be simplified by the observation that the vector $(1,1,...,1)^T$ is an eigenvector of $K$ whose eigenvalues is null and that $K$ is defined non-negative. Therefore:

$$\kappa_2(A) = \frac{\lambda_{\max}(M_0)}{\lambda_{\min}(M_0)} + \frac{\lambda_{\max}(K)}{\beta}, \quad (3.21)$$

which can be put in the following form:

$$\kappa_2(A) = 1 + \frac{\lambda_{\max}(M_0) - \lambda_{\min}(M_0)}{\lambda_{\min}(M_0)} + \frac{\lambda_{\max}(K)}{\beta}. \quad (3.22)$$

On the other hand, a bound for $\lambda_{\max}(A)$ can be obtained by the relation:

$$\lambda_{\max}(A) \leq \max_i \sum_i |k_{ij}|, \quad (3.23)$$

From (3.22)-(3.23) we see that the higher $\beta$ and the lower is the difference $\lambda_{\max}(M_0) - \lambda_{\min}(M_0)$ the better is the conditioning of matrix $A$ and consequently the stability of the linear system (3.16). As we have

$$\beta = \left(\frac{\mu}{(\tau)^2} + \frac{\gamma}{2\tau}\right). \quad (3.24)$$

it is clear that we can improve the stability by increasing $\mu$ and $\gamma$. So, by the Lax Equivalence Theorem [Hirsch(1988)], we can guarantee that the numerical scheme is convergent.

Although the gain of stability, we have to be careful with large values of $\beta$ because they can slowing down the convergence rate. Next, we show this effect in a quantitative way.

### 3.1.4 Terminating Criterion

The criteria found to stop the interactions for the system (3.16) are basically related to the (low) velocity of the snake in the neighborhood of the border.

In the first one [Cohen (1991)], we stop the interactive process when:

$$\left\|(X^n, Y^n) - (X^{n-1} - Y^{n-1})\right\| < \varepsilon, \quad (3.25)$$

for an error $\varepsilon$.

In the second one we postulate that close to the border the Mean Field Energy $E_l$ of the solution $c(s,t)$ has low variation, that is:

$$E_l(t) = \frac{\int_0^1 E_c ds}{\int_0^1 \left\|\frac{dc}{ds}\right\| ds}, \quad (3.26)$$

is such that:

$$|\Delta E_l| \equiv |E_l(t) - E_l(t-\tau)| < \varepsilon, \quad (3.27)$$

for an error $\varepsilon$.

The first criterion, called the *steady-state criterion*, is a stronger stability condition for the optimization scheme. The main problem of it is its sensibility to the oscillations due the elastic energy term and the inertia term (3.2). Due such oscillations more interactions could be necessary to be bellow the threshold in (3.25) which can bring performance problems.

The second criterion, called *steady-support* criterion, is proposed in [Leymarie-Levine (1993)]. The basis for this criterion is the observation that we are seeking for valleys and folds of the potential surface defined by (3.7). So, if the snake is close the valley we seek, the "mean field energy over the snake" given by (3.26) is expected to be less sensible to local motions of the snake than the first criterion.

### 4. Qualitative Analysis

The analysis of the numerical scheme presented in section 3.13 is a way of obtain relations between the parameters of the model. Another possibility for this is to find information in advance about the behavior of the continuum solution

Let's take a simple analysis of the snake system.

Firs notice that the equation (3.10) describes a system whose mechanical energy (potential plus kinetic) decrease in a rate given by (3.3). So, the corresponding point in the configuration space stops moving somewhere (steady-state). From the equation (3.10) we can show (bellow) that such a steady-state point is an extreme of the potential energy.

So, regardless the initial conditions and parameters, such simple analysis shows that an extreme of the potential energy is always achieved. Two questions naturally arise:

First: as the potential energy is not convex in general, how to find appropriate initial conditions and parameters to achieve exactly the local minimum we want?

Second: What about the rate in which the solution of the problem (3.10) goes to that local minimum?

In this and the next section we present an answer for the second question which links the convexity analysis of the energy and the dynamic formulation of snakes. In sections 7 we discuss the problem related to the first question by using the framework of the Hamiltonian formalism.

The result we are going to show is the narrow relation between the dynamic of the snake model and the eigenvalues of the Hessian of the Energy given by (3.8).

So, let's first put the equation (3.10) in the form:

$$\ddot{Q} = f(Q, \dot{Q}), \quad (4.1)$$

which is a ordinary non-linear second order equation.

Let's reduce the second order differential equation in a first order system by the change:

$$Q_1 = Q, \quad Q_2 = \dot{Q}, \quad (4.2)$$

So, the equation (3.10) becomes:

$$\begin{cases} \dot{Q}_1 = Q_2, \\ \dot{Q}_2 = f(Q_1, Q_2). \end{cases} \quad (4.3)$$

Let's call:

$$x = \begin{bmatrix} Q_1 \\ Q_2 \end{bmatrix}, \quad X_1 = Q_2, \quad X_2 = f(Q_1, Q_2), \quad (4.4)$$

so, the system (4.3) becomes

$$\dot{x} = X(x), \quad (4.5)$$

where $X(x) = (X_1, X_2)^T$. The system (4.5) is an autonomous first-order system [Sotomayor (1979)]. If the energy $E_p$ is $C^2$ then the field $X$ will be $C^1$ and the Cauchy problem:

$$\dot{x} = X(x), \quad x(0) = x_0, \quad (4.6)$$

has a unique solution.

From the dynamic system theory we know that the points that solve the equation:

$$X(x) = 0, \quad (4.7)$$

called singular (or critical) points, are fundamental for the qualitative characterization of the solutions of (4.5). A fundamental result emerges from the analysis of these points in the context of snakes. First we have to notice that these points are exactly the extremes of the energy $E_p$.

To show this let's call $\bar{x}$ one such solution of (4.7). So, by the definitions of $Q_1$ e $Q_2$ given in (4.2) we have:

$$\dot{Q} = 0, \quad (4.8)$$

for all $t$, so:

$$Q = \text{Const.} \quad (4.9)$$

If we take the snake model we see that such a point, which is a solution of the system (3.10), is also a solution of the equation:

$$\left(\frac{1}{2}\omega_1 M_1 + \frac{1}{2}\omega_2 M_2\right)Q = F(Q), \quad (4.10)$$

and so, is also an extreme of the energy functional (3.8) and might be the border we seek. Conversely if a point $\bar{x}$ is a border so $Q_2 = 0$ and it satisfies (4.10) then it is a singular point of *X*.

Therefore, the basic information of the topology of the phase space for (4.5) are related to the solutions of (4.10) which are in fact the points we are seeking. For the model (3.10) to be efficient, what has to be the relation between this topology and the border we seek?

The only possibility is that the border is an attractor, that is, a node or stable focus (figure 1.a, 1.c an 1.d, page 17). The other possibilities for *hyperbolic singular points* (real part non null) are also represented in the figure 1 and it is quite clear to understand the difficulties inherent, that is, we always have *unstable subspaces* in which the solution goes far from the desired point (figure 1.b, 1.e).

The *non-hyperbolic* case (some eigenvalues with real part null) has bad consequences. The unstable focus is an example of this situation (figure 2). Besides having subspaces in which we do not have convergence to the solution, there is another problem related to the *structural stability* of the system. In fact the system in this case is structurally unstable [Sotomayor (1979)] which implies numerical problems due bifurcations that might happens. So, it is an undesirable situation.

Therefore, as we want the border we seek to be an attractor, we have to chose the parameters $(\mu, \gamma, \omega_1, \omega_2)$ in a way that the eigenvalues of *DX* have real part strictly negative [Sotomayor (1979)] in the solution $\bar{x}$ of (4.10) that we want This is the central result of this section which can be summarized as follows:

*Claim 1: The parameters of the model have to be chosen in a way that the border we seek is an attractor or stable focus in the topology of the phase space of (4.5)*

In the next section we will demonstrate that a sufficient condition for having this is that the energy $E_p$ has a local minimum in the border we seek.

## 5 Convexity and Attractors

We prove in this section the following statement:

*Claim 2: A singular point x of X is an attractor or stable focus if the energy (3.8) is strictly convex in a neighborhood of x, that means, if x is a local minimum of $E_p$.*

To demonstrate this we have to analyze the eigenvalues of the matrix *DX* in a singular point *x* supposing that the potential energy is convex in a neighborhood $V_x$ of *x*. If these eigenvalues have real part non null then we can use the Hartman's theorem to linearize the field X in $V_x$ [Sotomayor (1979)]. Therefore, from the linear dynamics system theory, if all the eigenvalues of DX(x) have the real part strictly negative we will have an node or stable focus and the statement is demonstrated.

From (5.1)-(5.3) we see that the matrix *DX* has the form:

$$DX(x) = \begin{pmatrix} 0 & I_{NxN} \\ -\frac{1}{\mu}M_0^{-1}D^2(E_p) & -\frac{\gamma}{\mu}I_{NxN} \end{pmatrix}, \quad (5.1)$$

where $I_{NxN}$ is the identity *NxN* matrix, $D^2(E_p)$ is the Hessian matrix of the potential energy (3.8) and $M_0^{-1}$ is the inverse of $M_0$ which is defined in (3.2).

The eigenvalues/eigenvectors equation for *DX(x)* is:

$$DX \cdot v = \sigma v, \quad (5.2)$$

where $(\cdot)$ means matrix/vector multiplication.

By using (5.1) we can rewrite (5.2) in the form:

$$\begin{cases} v_2 = \sigma v_1 \\ -\frac{1}{\mu}M_0^{-1}D^2(E_p) \cdot v_1 - \frac{\gamma}{\mu}I \cdot v_2 - \sigma v_2 = 0 \end{cases} \quad (5.3)$$

By using the first equation above we can rewrite the second one only in function of $v_1$ to obtain:

$$\frac{1}{\mu}M_0^{-1}D^2(E_p) \cdot v_1 + \sigma\frac{\gamma}{\mu}I \cdot v_1 + \sigma^2 v_1 = 0. \quad (5.4)$$

Remember that we are only interested in the signal of the eigenvalue $\sigma_i$, $i = 1,...,2N$ supposing that the eigenvalues $\alpha_i$, $i = 1,...,2N$ of the Hessian $D^2(E_p)$ are strictly positive. First, we have to notice that the equation (5.4) can be put in the form:

$$D^2(E_p) \cdot v_1 = \beta M_0 \cdot v_1 = 0, \quad (5.5)$$

where

$$\beta = -(\sigma\gamma + \mu\sigma^2) \quad (5.6).$$

So, we have a generalized eigenproblem very known in the field of vibration mode superposition analysis for mechanical structures. It can be shown [Bathe-Wilson (1976)] that if $M_0$ and $D^2(E_p)$ are positive-defined the eigenvectors $\phi_i, i=1,...,N$, corresponding to (5.5) are $M_0$- and $D^2(E_p)$-orthogonal; that means:

$$\phi_i^T M_0 \phi_j = \delta_{ij}, \quad \phi_i^T D^2(E_p)\phi_j = \beta_i \delta_{ij}. \quad (5.7)$$

By using these relations, it is straightforward to show that equation (5.5) is equivalent to:

$$\beta_i + \sigma\gamma + \mu\sigma^2 = 0, \ i = 1,...,N, \quad (5.8)$$

whose solutions are given by:

$$\sigma_i^\pm = \frac{-\gamma \pm \sqrt{\Delta_i}}{2\mu}, \quad i = 1,..,N.,$$

(5.9)

where

$$\Delta_i = \gamma^2 - 4\mu\beta_i. \quad (5.10)$$

For *x* to be a attractor we need that:

$$4\mu\beta_i > 0, \ i = 1,...,N. \quad (5.11)$$

As $M_0$ represents a norm, a sufficient condition for (5.11) is that $D^2(E_p)$ is positive defined according to the properties of the generalized eigenproblem (5.5). So, if the border we seek is a local minimum we have the above condition. Then, the Claim 2 is demonstrated.

**6. Discussion**

Before analyzing the result (5.11) it is convenient to take some considerations.

First, it is interesting to consider explicitly the fact that the potential energy (3.8) is dependent of not only the control points vector Q but also from the parameters $\omega_1$ and $\omega_2$. So, following the classical nomenclature, $E_P$ is a family of functions (potentials) of the form:

$$E_P : \mathfrak{R}^{2N+2} \times \mathfrak{R}^2 \to \mathfrak{R}; \quad (6.1)$$
$$E_P = E_P(Q;\omega_1,\omega_2).$$

In this context, the result (5.20) is precisely stated as:

$$4\mu\alpha_i(Q;\omega_1,\omega_2) > 0, \ i = 0,1,...,N. \quad (6.2)$$

as the eigenvalues of $D^2(E_P)$ will be dependent of $Q$ and $(\omega_1,\omega_2)$ also.

If $E_P$ is strictly convex in a region $R'$ a direct consequence is that $E_P$ is equivalent (in topological sense [Poston-Stewart (1987)]) to

$$E_P + f : R' \times \mathfrak{R}^2 \to \mathfrak{R}; \quad (6.3)$$
$$f = f(Q;\omega_1,\omega_2).$$

where f is a perturbation.

As a consequence [Poston-Stewart (1987)] the extremum of (3.8) moves as nice function of the parameters $(\omega_1,\omega_2)$.

In the dynamical point of view, such *structural stability* of $E_P$ implies that the field X in (4.5) and that one correspondent to (Ep+f) are topologically equivalent [Sotomayor (1979)] and so the qualitative behavior of the solutions are the same.

If $E_P$ is nonconvex, the lack of structural stability corresponding implies in numerical instability and gives splits and discontinuities in the final contour [Davatzikos-Prince (1996)].

The main problem here is to specify a region $R'$ in which $E_P$ is strictly convex. In the case of the model (2.6) this region is specified by (2.21).

In our case we have to consider (3.24) and (5.9) also because our model involves the velocity space. The expression (5.9) is an important result because it links the two main elements of the snake model (3.10): the potential energy (through the signal of $\beta_i$) and the dynamic (parameters $\mu$ and $\gamma$).

The first consequence of (5.9) is that it is not possible to have "critical dumping" in all "modes" $\phi_i, i=1,...,N$ in (5.7), because it is not possible to satisfy the equation (5.20) for all i and j.

Also, the result (5.9) can be viewed as a complement of the result (3.23). In that case, we notice that the bigger $\gamma$ and $\mu$ the better is the numerical stability. But, now we see by (5.9) that larger values of these quantities could make $\Delta_i$ more negative and so the oscillatory term bigger. Such effect makes the convergence gets worse.

It is important to stress that if (5.11) is true in a singular point *x* than, by Hartman's Theorem, there is a neighborhood $V_x$ of x in which the solution of (3.10) is always attracted to *x* independent of the initial conditions $x_0 \in V_x$ that we choose. The main point now is how to precise $V_x$.

The energy (3.8) is not enough for this analysis because it do not incorporate the velocity space. Such observation points forward to the necessity of extend the convexity analysis to the velocity space.

To be precise, we have to find a physical quantity whose convexity analysis is determinant of the dynamic and which incorporate the velocity space $Q_2$. This quantity is exactly the mechanical energy $\left(T + E_p\right)$ of the system whose properties of interest arises naturally from the *Hamiltonian Formulation* of the snake model which will be presented in the next section.

### 7. Hamiltonian Equations of Motion

The equations of motion of a system in the classical mechanics can be established by the Lagrangean formulation or by the Hamiltonian formulation. Although equivalents, the usefulness of the Hamiltonian viewpoint lies in providing a framework for theoretical extensions in areas of physics such as Hamilton-Jacobi Theory and perturbation approaches [Goldstein (1980)].

In the case of snakes models the Hamiltonian Formulation gives a natural way of extending the convexity analysis presented in section (2.1) to the dynamic model given by (3.10).

#### 7.1 Hamilton's Formulation of Snakes

In the Lagrangian viewpoint a system with *n* independent degrees of freedom is a problem in *n* independent variables $q_i(t)$, and $\dot{q}_i$ appears only as a shorthand for the derivative of $q_i$.

The Hamiltonian formulation is based on a fundamentally different picture. We seek to describe the motion in terms of first-order equations of motion. To achieve this goal let's first define the so called generalized or *conjugate momenta* $p_i$ as [Goldstein (1980)]:

$$P = \frac{\partial L(Q, \dot{Q}, t)}{\partial \dot{Q}}. \quad (7.1)$$

The quantities *(Q,P)* are known as the *canonical variables.*

The transformation from $\left(Q, \dot{Q}, t\right)$ to $\left(Q, P, t\right)$ is accomplished by the *Hamiltonian* of the system defined by [Goldstein (1980)]:

$$H(Q,P,t) = \dot{Q}^T P - L(Q, \dot{Q}, t). \quad (7.2)$$

Considered as a function of *Q, P* and *t* only, the differential of *H* is given by:

$$dH = \left(\frac{\partial H}{\partial Q}\right)^T dQ + \left(\frac{\partial H}{\partial P}\right)^T dP + \frac{\partial H}{\partial t} dt, \quad (7.3)$$

but from the defining equation (7.2) we can also write:

$$dH = \dot{Q}^T dP + P^T d\dot{Q} - \left(\frac{\partial L}{\partial \dot{Q}}\right)^T d\dot{Q} - \left(\frac{\partial L}{\partial Q}\right)^T dQ - \frac{\partial L}{\partial t} dt, \quad (7.4)$$

Using the definition of generalized momentum we can see that the terms in $d\dot{Q}$ cancel. From Lagrange's Equations it follows that:

$$\frac{\partial L}{\partial Q} = \dot{P} + \gamma M_0 \dot{Q}, \quad (7.5)$$

Equation (7.4) therefore reduces to:

$$dH = \dot{Q}^T dP - \dot{P}^T dQ - \frac{\partial L}{\partial t} dt, \quad (7.6)$$

Comparing this equation with (7.3) we find the following set of *2n+1* equations:

$$\begin{cases} \dot{Q} = \frac{\partial H}{\partial P}, \\ \dot{P} = -\frac{\partial H}{\partial Q} - \frac{\gamma}{\mu} P, \quad (7.7) \\ \frac{\partial L}{\partial t} = -\frac{\partial H}{\partial t}. \end{cases}$$

If γ=0, equations (7.7) are known as the *canonical equations of Hamilton;* they constitute the desired set of *2n+1* first order equations of motions replacing the Lagrange's equations.

In our case, we do not have explicit dependency of time, so the last equation is trivially satisfied.

Using the conjugate momenta defined in (7.1) we find that:

$$P \equiv (p_1, p_2, \ldots, p_n)^T = \mu M_0 \dot{Q}. \quad (7.8)$$

Therefore, we can rewrite $H$ in terms of the canonical variables $(q, p)$:

$$H(q,p) = \frac{1}{2\mu} P^T M_0^{-1} P + Q^T K Q + F(Q), \quad (7.9)$$

which is exactly the mechanical energy of the system, that is:

$$H(q,p) = T + E_p. \quad (7.10)$$

First we have notice by using (7.10) that a singular point of $X$ in (4.5) is also a singular point of (7.7) and conversely. So, the points we seek are also singular points of the dynamic system (7.7). Due (7.8), as they are of the form $(Q_1, 0)$ in the $(Q, \dot{Q})$ space they will be of the form $(Q, 0)$ in the $(Q, P)$ also.

We have also the following properties for the Hamiltonian.

*Property 1.* If $\nabla E_p(x_b) = 0$ then $\nabla H(x_b, 0) = 0$.

Dem.

$$\nabla H(x_b) = \left(\frac{\partial H}{\partial Q}, \frac{\partial H}{\partial P}\right)\bigg|_{x_b} = \left(\nabla E, \frac{1}{\mu} M_0^{-1} P\right)_{(Q_b, 0)} = (0, 0)$$

*Property 2.* If the potential energy $E_p$ in (3.8) is strictly convex in a neighborhood $V_{x_b}$ so the Hamiltonian (7.10) is also strictly convex in $V_{x_b}$ if the matrix $M_0^{-1}$ is positive defined.

Dem. Supposing that $D^2 E_p(x_b)$ is positive defined. The Hessian of H is:

$$D^2 H(x_b) = \begin{pmatrix} \frac{\partial^2 H}{\partial Q^2} & \frac{\partial^2 H}{\partial P \partial Q} \\ \frac{\partial^2 H}{\partial Q \partial P} & \frac{\partial^2 H}{\partial P^2} \end{pmatrix}\Bigg|_{x_b} = \begin{pmatrix} D^2 E_p(x_b) & 0 \\ 0 & \frac{1}{\mu} M_0^{-1} \end{pmatrix},$$

$$(7.11)$$

which is in positive defined in $V_{x_b}$ if the $D^2 E_p$ and the constant matrix $M_0^{-1}$ are positive defined.

From the properties 1 and 2 we see that if $x_b$ is a local minimum of the $E_p$ and $M_0^{-1}$ is positive defined we have a local minimum of the Hamiltonian in $(Q = x_b, P = 0)$.

It is natural to seek for the results (5.9)-(5.11) in the *(Q,P)* space. To do this we have to use the Hamilton's equations (7.7) to define the following field Y:

$$Y(Q,P) = \left(\frac{\partial H}{\partial P}, -\frac{\partial H}{\partial Q} - \frac{\gamma}{\mu} P\right). (7.12)$$

First we have notice that if $x_b = (Q_b, 0)$ is a singular point for the field X in (4.5) so it will be a singular point for Y in the *(Q,P)* space because:

$$\frac{\partial H(Q_b, 0)}{\partial P} = 0, \quad \frac{\partial H(Q_b, 0)}{\partial Q} = \nabla E_p(Q_b) = 0.$$

For $x_b = (Q_b, 0)$ to be an attractor we have to study *DY* given by:

$$DY(x_b) =$$

$$\begin{pmatrix} \dfrac{\partial^2 H}{\partial Q \partial P} & \dfrac{\partial^2 H}{\partial P^2} \\ -\dfrac{\partial^2 H}{\partial Q^2} & -\dfrac{\partial^2 H}{\partial Q \partial P} - \dfrac{\gamma}{\mu}I \end{pmatrix}\Bigg|_{x_b} =$$

$$\begin{pmatrix} 0 & \dfrac{1}{\mu}M_0^{-1} \\ -D^2 E_p(x_b) & -\dfrac{\gamma}{\mu}I \end{pmatrix}$$

(7.13).

Therefore, eigenvalue equation for $DY(x_b)$ is:

$$DY(x_b)\begin{pmatrix} v_1 \\ v_2 \end{pmatrix} = \sigma\begin{pmatrix} v_1 \\ v_2 \end{pmatrix} \Leftrightarrow$$

$$\begin{cases} \dfrac{1}{\mu}M_0^{-1} \cdot v_2 = \sigma v_1, \\ -D^2 E_p(x_b) \cdot v_1 - \dfrac{\gamma}{\mu}I \cdot v_2 = \sigma v_2 \end{cases} \quad (7.14).$$

By isolating $v_2$ in the first equation and substituting in the second equation we will find an equation equivalent to (5.8)-(5.9).

So, like in (5.11), if $\alpha_j > 0$ (local minimum) we have that $(x_b, 0)$ is an attractor in the (Q,P) space.

### 7.2 Snake and the Hamiltonian

First, let's notice that:

$$(-\nabla H) \cdot Y = \dfrac{\gamma}{\mu}\|\dot{Q}\|_2^2 > 0, \text{ if } \dot{Q} \neq 0. \quad (7.15)$$

From this result we conclude that the solution of the Cauchy problem for the Hamilton's Equations (7.7) with $\gamma \neq 0$ has always a component in a direction that minimizes the Hamiltonian. If $\gamma = 0$, the Hamiltonian is conserved which is a very known result in classical mechanics [Goldstein (1980)].

So, let's take a Cauchy problem for (7.7). If we guarantee that the solution $\varphi(t)$ is always inside the region $R'$ of convexity of $E_p$ (and of $H$ for property 2) then through (7.15) the only possibility for the solution would be to go to the local minimum of the Hamiltonian in $R'$ which is the only local minimum of $E_p$ in $R'$. But, how to guarantee that:

$$\varphi(t) = (Q(t), P(t)) \text{ is such that } Q(t) \in R', \forall t ?$$

To solve this let's take the initial conditions $(Q_0, P_0)$ such that $Q_0 \in R'$ and:

$$H(Q_0, P_0) \leq \min E_P|_{boudary}, \quad (7.16)$$

where $\min E_p|_{boudary}$ means the minimum of $E_p$ in the boundary of $R'$. Inequality (7.15) implies that:

$$T(P_0) + E_p(Q_0) \leq \min E_P|_{boudary}. \quad (7.17)$$

As the system is losing mechanical energy according (7.15) the only possibility for the $\varphi(t)$ is such that:

$$T(P(t)) + E_p(Q(t)) \leq \min E_P|_{boudary}$$
$$\Leftrightarrow$$
$$E_p(Q(t)) \leq \min E_P|_{boudary}, \quad (7.18)$$

as $T(P) \geq 0$ for all $P$. As $E_p$ is strictly convex in $R'$ the inequality (7.18) means that:

$$\varphi(t) = (Q(t), P(t)) \text{ is such that } Q(t) \in R', \forall t.$$

Figure 3 helps to understanding the result. In this figure the horizontal lines represent the Hamiltonian for two choices of $(Q_0, P_0)$. In the first one, the corresponding value of $H$ do not satisfies (7.16). In this case, the system can go away the global minimum due to the inertia (velocity). However, if $H$ satisfies (7.16) for $(Q_0, P_0)$ the system has to has to go to the (global) minimum due (7.15) and the convexity of $E_p$.

Now, we can state our final conclusion:

Claim 3:

"Having found a region $R'$ in which the $E_p$ is strictly convex, the condition (7.16) for the initial conditions is sufficient to guarantee the convexity of the solution $\varphi(t)$ to the desired point."

### 8. Conclusions and Future Works

The main conclusions of this work can be summarized by the Claim 2 of section 5 and the result (5.9).

The result (5.9) shows quantitatively how the dynamic parameters govern the rate of convergence of the numerical method. Although it was expected that bigger values for μ and γ could slow down that rate [Leymarie-Levine(1993)], a precise characterization of this was not found in the literature.

The Claim 2 (section 5) gives the exactly relation between the dynamic snake model and the potential energy (3.8): the (global) minimum of the potential energy has to be an attractor in the phase space of the dynamic model.

The Claim 3 in the last section generalizes the convexity analysis presented in [Davatzikos-Prince (1996)] by incorporating the velocity space. The practical utility of this result depends on the possibility of establishing the condition (7.16).

To do this, we have to develop the convexity analysis presented in [Davatzikos-Prince (1996)] in the context of the dynamic snake model. Next, we have to analyze the possibilities of avoiding the main difficulty expected for establishing condition (7.16): the computational cost.

We can also substitute the damping matrix $\gamma M_0$ by the Caughey series [Bathe-Wilson(1976)]:

$$C = M \sum_{k=0}^{p-1} \left[ M_0^{-1} K \right]^k, \quad (8.1)$$

as a way of getting critical damping in more modes $\phi_i$ of (5.7). The expression (8.1) has the advantage of maintaining the same transformation to diagonalise the damping matrix [Bathe-Wilson(1976)].

Another possible direction for this work is to analyze tracking models, like the work found in [Peterfreund (1999)], with the framework presented here. Probably, that is the better context to analyze the practical consequences of this work.

**Acknowledgments**

We would like to thanks CNPq for the financial support.

**References**

[Bartels at al. (1987)] R. H. Bartels, J. C. Beatty, and B. A. Barsky, "An Introduction to Splines for use in Computer Graphics & Geometric Modeling", Morgan Kaufmann Publishers, INC, Los Altos, California, 1987.

[Bathe-Wilson(1976)] K. J. Bathe and E. L. Wilson, "Numerical Methods in Finite Element Analysis", Prentice-Hall, INC, Englewood Cliffs, New Jersey, 1976.

[Black-Yuille (1992)] Active Vision, MIT Press, 1992.

[Chapra-Canale (1988)] C. Chapra and R.P. Canale. "Numerical Methods for Engineers." MacGraw-Hill International Editions, 1988.

[Cohen (1991)] L. D. Cohen, "On Active Contour Models and Balloons", *CVGIP: Image Understanding*, No. 2, March 1991, 211-218.

[Cohen-Cohen (1993)] "Finite-Element Methods for Active Contour Models and Balloons for 2-D and 3-D Images", *IEEE Trans. on Pattern Analysis and Mach. Intelligence,* 15 (11), November 1993.

[Curwen-Blake (1993)] R. Curwen and Blake, " Dynamic Contours: Real-time Active Splines", in *Active Vision*, edited by A. Black and A.Yuille, MIT Press(1993).

[Davatzikos-Prince (1996)] C. Davatzikos and J. L. Prince, "Convexity Analysis of Active Contour Problems", *Proceedings of CVPR*, San Francisco, June 17--20, 1996. Web Site: http://iacl.ece.jhu.edu/~prince/jlp_pubs.html

[Dubrovin at al. (1984)] B. A. Dubrovin, A. T. Fomenko, and S. P. Novikov, "Modern Geometry - Methods and Applications. Part 1. The Geometry of Surfaces, Transformation Groups, and Fields", Springer-Verlag, 1984.

[Fisker-Carstensen (1998)] "On parameter estimation in deformable models", 14[th] International Conference on Pattern Recognition, Vol. 1, pp. 762-766, August 1998.

[Goldstein (1980)] H. Goldstein, *Classical Mechanics,* second edition, 1980.

[Golub-Loan(1985)] G. H. Golub and C. F. Van Loan, *Matrix Computations,* The Joans Hopkins University Press, 1985.

[Grenander-Szego (1958)] U. Grenander and G. Szego, "Toeplitz Forms and their Applications", University of California Press, 1958.

[Gunn-Nixon (1997)] S. R. Gunn and M. S. Nixon, "A Robust Snake Implementation; A Dual Active Contour", *IEEE Trans. on Pattern Analysis and Mach. Intelligence,* 19 (1), January 1997.

[Hirsch (1988)] C. Hirsch, "Numerical Computation of Internal and External Flows: Fundamentals of Numerical Discretization", Vol. 1, John Wiley & Sons, 1988.

[Kass at al. (1987)] M. Kass, A. Witkin and D. Terzopoulos, "Snakes: Active contour models", *Proc. First Int. Conf. Comput. Vision*, pp. 259-268, London, 1987.

[Leymarie-Levine (1993)] F. Leymarie and M. D. Levine, " Tracking Deformable Objects in the Plane Using an Active Contour Model", *IEEE Trans. on Pattern Analysis and Mach. Intelligence,* 15 (6), June 1993.

[Nayfeh-Mook (1979)] A. H. Nayfeh and D. T. Mook, "Nonlinear Oscillations", *John Wiley & Sons*, a Wiley-Interscience Publication, 1979.

[Peterfreund (1999)] N. Peterfreund, "Robust Tracking of Position and Velocity with Kalman Snakes", *IEEE Trans. on Pattern Analysis and Mach. Intelligence,* 21 (6), June 1999.

[Poston-Stewart 91987)] T. Poston and I. Stewart, "Catastrophe Theory and its Applications", Pitman Publishing Limited, London, 1987.

[Qin-Terzopoulos (1996)] "D-NURBS: A Physics-Based Framework for Geometric Design", *IEEE Trans. on Visualization and Comp. Graphics,* Vol. 2, No. 1, March 1996.

[Scalaroff-Pentland (1995)] "Modal Matching for Correspondence and Recognition", *IEEE Trans. on Pattern Analysis and Mach. Intelligence,* 17 (6), June 1995

[Sotomayor (1979)] J. Sotomayor, "Lições de Equações Diferenciais Ordinárias", *Projeto Euclides*, Gráfica Editora Hamburgo Ltda, São Paulo, 1979.

[Xu at al. (1994)] "Robust Active Contours with Insensitive Parameters", Pattern Recognition, Vol. 27, No. 7, pp. 879-884, 1994.

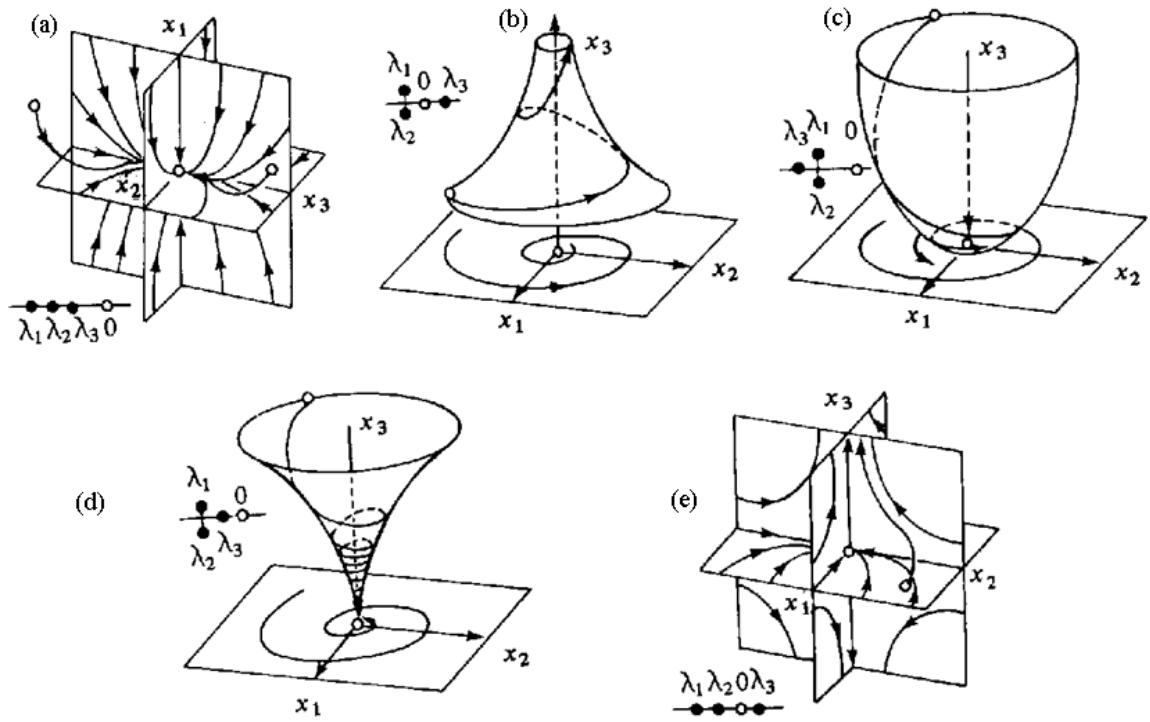

Figure 1. Hyperbolic ($\dot{x}=Ax$) Systems in $R^3$. The eigenvalues of A are $\lambda_1 \lambda_2 \lambda_3$.

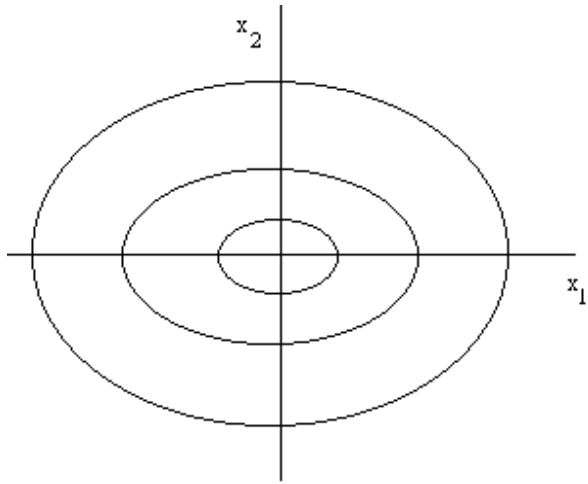

Figure 2. Unstable focus for a 2D system. Eigenvalues $\lambda_1 = ai$ and $\lambda_2 = bi$

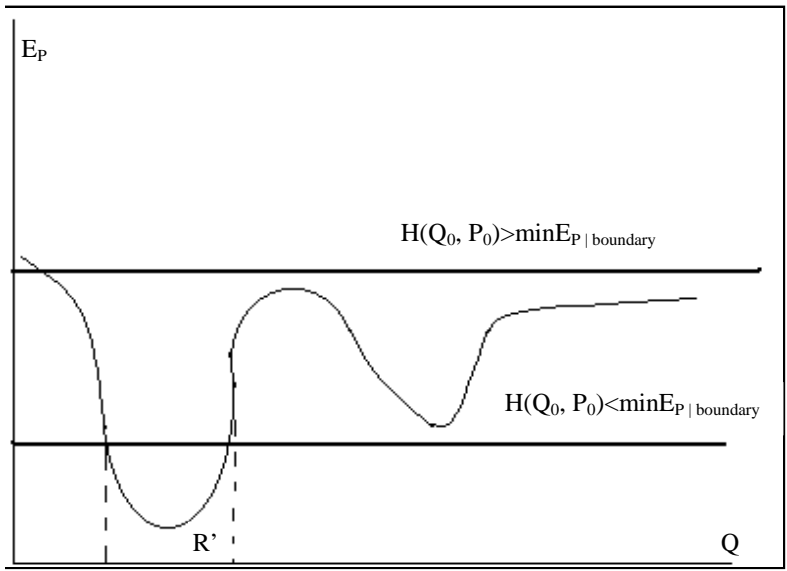

Figure 3. Two choices of initial conditions $(Q_0, P_0)$. In H do not satisfies (7.16) the system can go to regions where $E_p$ is not convex. But, if H satisfies (7.16) the system do not go away from R' and the convergence is guaranteed.